\documentclass[10pt,twocolumn,letterpaper]{article}

\usepackage{cvpr}              

\usepackage[pdftex]{graphicx}
\usepackage[numbers]{natbib}
\usepackage{graphicx}
\usepackage{amsmath}
\usepackage{amssymb}
\usepackage{booktabs}
\usepackage{hhline}
\usepackage{multirow}
\usepackage{multicol}
\usepackage{microtype}      
\usepackage{xcolor}         
\usepackage{hhline}
\usepackage{bbm}
\usepackage{boldline}
\usepackage{graphicx}
\usepackage{amssymb}
\usepackage{pifont}
\usepackage{soul}
\usepackage[accsupp]{axessibility}
\usepackage{float} 
\newcommand{\cmark}{\ding{51}}%
\usepackage{hyperref}
\hypersetup{
    colorlinks=true,
    linkcolor=blue,
    filecolor=magenta,      
    urlcolor=cyan,
}
\usepackage[capitalize]{cleveref}
\crefname{section}{Sec.}{Secs.}
\Crefname{section}{Section}{Sections}
\Crefname{table}{Table}{Tables}
\crefname{table}{Tab.}{Tabs.}

\def\cvprPaperID{7353} 
\def\confName{CVPR}
\def\confYear{2022}

\newcommand{\Skip}[1]
{
}
\begin{document}

\title{Task Discrepancy Maximization for Fine-grained Few-Shot Classification}

\author{
SuBeen Lee, WonJun Moon, Jae-Pil Heo\thanks{Corresponding author}\\
Sungkyunkwan University\\
{\tt\small \{leesb7426, wjun0830, jaepilheo\}@skku.edu}
}
\maketitle
\begin{abstract}
Recognizing discriminative details such as eyes and beaks is important for distinguishing fine-grained classes since they have similar overall appearances.
In this regard, we introduce Task Discrepancy Maximization (TDM), a simple module for fine-grained few-shot classification.
Our objective is to localize the class-wise discriminative regions by highlighting channels encoding distinct information of the class. 
Specifically, TDM learns task-specific channel weights based on two novel components: Support Attention Module (SAM) and Query Attention Module (QAM).
SAM produces a support weight to represent channel-wise discriminative power for each class.
Still, since the SAM is basically only based on the labeled support sets, it can be vulnerable to bias toward such support set.
Therefore, we propose QAM which complements SAM by yielding a query weight that grants more weight to object-relevant channels for a given query image. 
By combining these two weights, a class-wise task-specific channel weight is defined.
The weights are then applied to produce task-adaptive feature maps more focusing on the discriminative details.
Our experiments validate the effectiveness of TDM and its complementary benefits with prior methods in fine-grained few-shot classification.
\footnote{Our code is available at \href{https://github.com/leesb7426/CVPR2022-Task-Discrepancy-Maximization-for-Fine-grained-Few-Shot-Classification}{https://github.com/leesb7426/CVPR2022-Task-Discrepancy-Maximization-for-Fine-grained-Few-Shot-Classification.}}
\end{abstract}

\section{Introduction}
\label{sec:intro}
\begin{figure}[t!]
    \centering
    \includegraphics[width=\columnwidth]{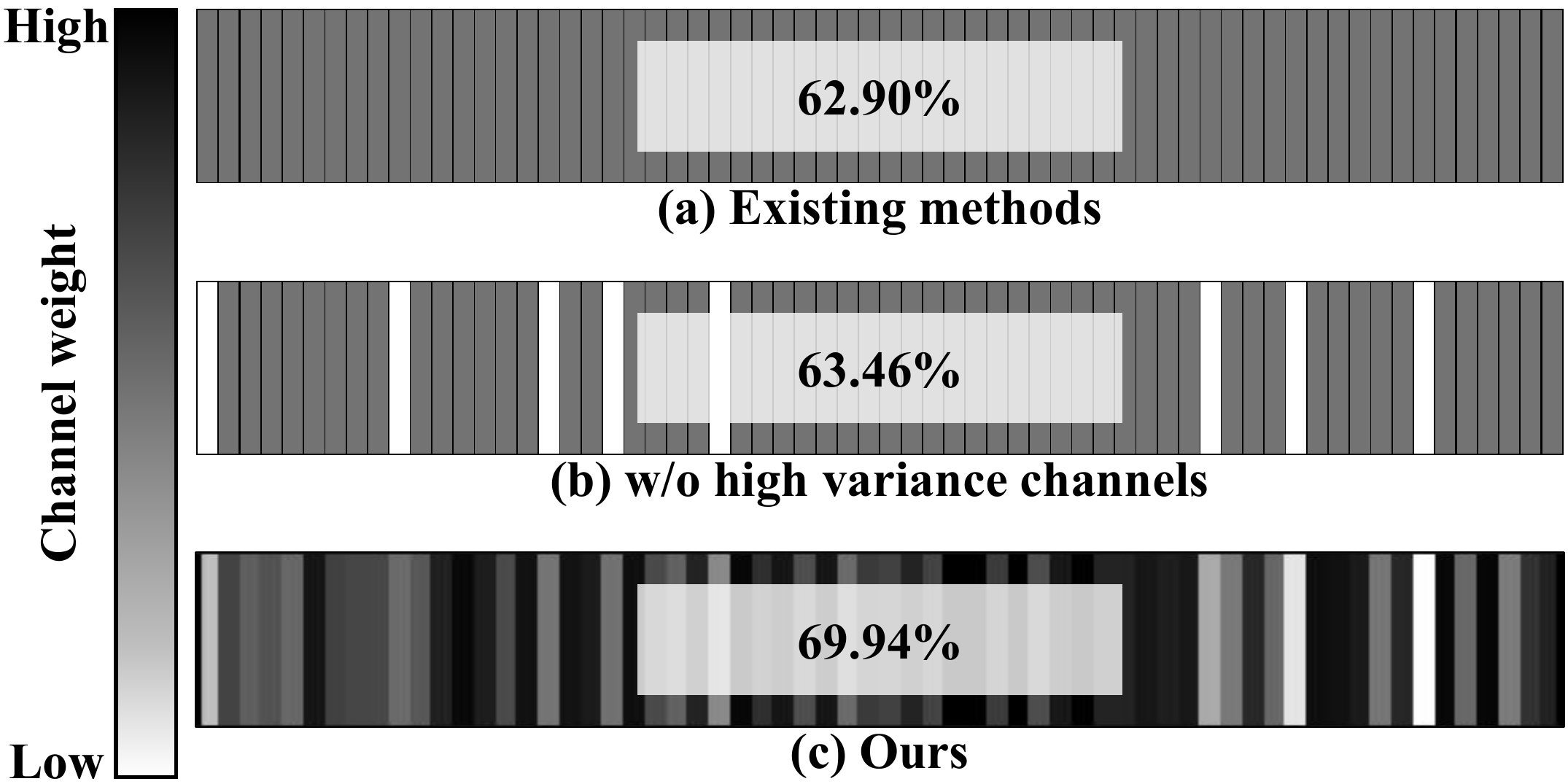}
    \vspace{-0.6cm}
    \caption{Effect of the channel weight in the CUB dataset. (a) Existing methods treat channels of feature maps equally. In such case, high variance channels within a class highly likely to disturb the classification task, where channel variance represents the channel-wise variances of feature maps of the same class -- intuitively, the instances of the same class having similar features at a channel lead a low channel variance for the corresponding channel. This is mainly because it is hard to make a consensus among features for classification criteria. (b) However, in fine-grained datasets, simply removing high variance channels for each class shows marginal improvements. It is because classes share similar features, e.g., feather, and wings in CUB dataset, and thus channels with a low variance may not be discriminative. Therefore, in fine-grained datasets, we should grant different weights to channels depending on whether each channel reflects distinct characteristics. (c) TDM produces per-class channel weight by discovering discriminative channels for each class in the episode. Note that, the numbers in boxes are classification accuracies.}
    \label{fig:motivation}
    \vspace{-0.5cm}
\end{figure}
With the advancement of deep learning, it has achieved remarkable performance beyond humans in various downstream tasks~\cite{deng2009imagenet, he2016deep}. 
However, there is a strong assumption that numerous labeled images should exist to achieve such performance.
If the number of labeled images is insufficient, it shows drastic degradation in the performance\cite{vinyals2016matching, finn2017model, chen2019closer}.
To resolve such degradation from a shortage of labeled images and reduce the cost of labeling, the computer vision community recently paid more attention to few-shot classification\cite{finn2017model, snell2017prototypical, vinyals2016matching}.
Briefly, the goal of few-shot classification is to train a model with high adaptability to novel classes.
To achieve this goal, the episodic learning strategy is mainly used, where each episode consists of sampled categories from the dataset. 
Furthermore, each class has a support set for training and a query set for evaluation.

The stream of metric-based learning is a promising direction for the few-shot classification.
These methods\cite{vinyals2016matching, snell2017prototypical, khrulkov2020hyperbolic,sung2018learning} learn a deep representation with a predefined metric or online-trained metric.
Specifically, the inference for a query is performed based on the distances among support and query sets under such metric.

However, the features of a novel class extracted by a model trained on the base classes hardly form a tight cluster, since the feature extractor is highly sensitive and activates the semantically discriminative variations in the distribution of base classes\cite{zhang2021prototype, roady2020difsim}.
To alleviate this, recent methods utilize primitive knowledge\cite{zhang2021prototype, li2020boosting} or propose task-dynamic feature alignment strategies\cite{kang2021relational, xu2021learning, doersch2020crosstransformers, simon2020adaptive, wertheimer2021few, ye2020few, hou2019cross}. 
Among two strategies, task-dynamic feature alignment methods are being spotlighted.
The task-dynamic feature alignment methods can be further divided into two main streams: spatial alignment and channel alignment. 
The spatial alignment methods\cite{hou2019cross, doersch2020crosstransformers, xu2021learning, kang2021relational, wertheimer2021few, wertheimer2021few} aim to resolve the spatial mismatch between key features on the feature maps of different instances.
On the other hand, the channel alignment methods\cite{xu2021learning, kang2021relational, ye2020few, simon2020adaptive} modify feature maps to better represent the semantic features for novel classes.

Although these alignment methods are shown to be effective on the general few-shot classification task, they achieved insignificant gains for fine-grained datasets.
This is mainly because they only focus on exploiting features that describe novel objects, which may not be discriminative in such tasks.
Indeed, localizing discriminative details is important in fine-grained classification, since categories share similar overall appearances~\cite{ge2019weakly, liu2020filtration, ding2019selective, zheng2019looking}.
Therefore, distinct clues for each category also should be discovered for fine-grained few-shot classification.
In \cref{fig:motivation} (c), we verify that localizing discriminatory details of the object through channel weights is effective for fine-grained few-shot task.

\begin{figure}[t!]
    \centering
    \includegraphics[width=\columnwidth]{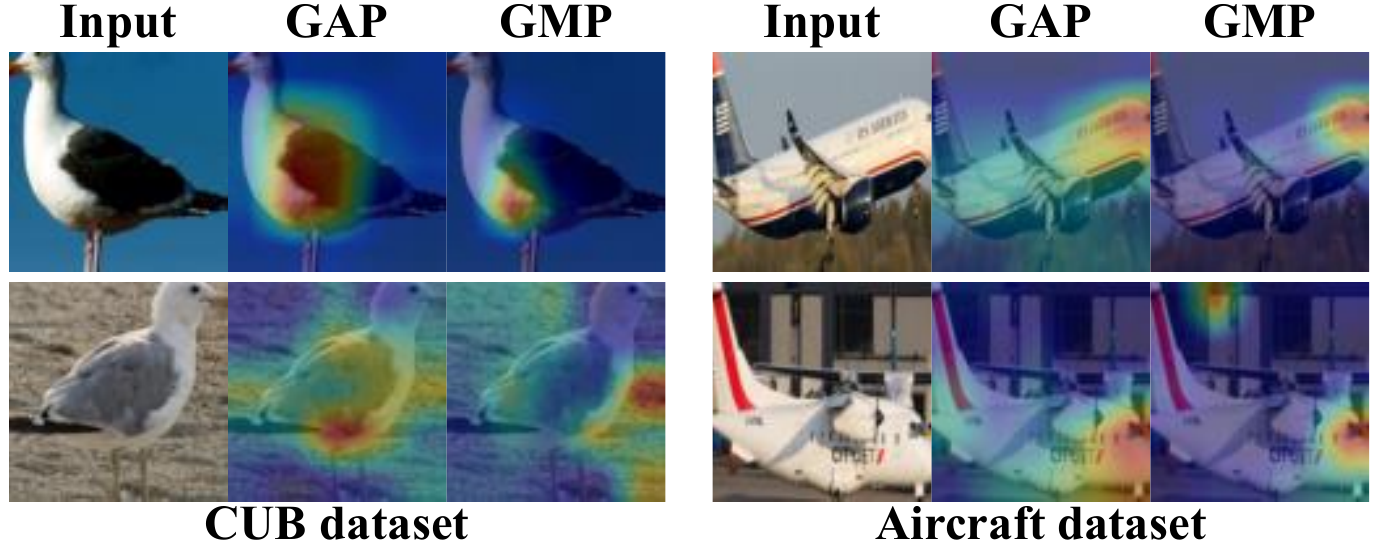}
    \vspace{-0.5cm}
    \caption{
    Visualization of pooling results.
    Each column shows the locations where each pooling method focuses on the image. 
    The second and third columns visualize the results of average pooling and the max pooling, respectively.
    GAP tends to concentrate on the object parts in the images, while GMP is often out of focus.
    }
    \vspace{-0.4cm}
    \label{fig:pooling}
\end{figure}

In this context, we introduce novel Task Discrepancy Maximization (TDM), a module that localizes discriminative regions by weighting channels per class. 
TDM highlights the channels that represent discriminative regions and restrains the contributions of other channels based on class-wise channel weight.
Specifically, TDM is composed of two components: Support Attention Module (SAM) and Query Attention Module (QAM).
Given a support set, SAM outputs a support weight per class that presents high activations on discriminative channels. 
On the other hand, QAM is fed with the query set to produce a query weight per instance. 
The query weight is to highlight the object-relevant channels.
To infer these weights, the relation between each feature map and the average channel pooled features are considered.
Note that, the channel pooled average feature map has the spatial information of the object\cite{woo2018cbam, li2020spatial} as described in \cref{fig:pooling}.
Therefore, channels are highly likely to represent objects when they are similar to spatially averaged feature map.
By combining two weights computed from our sub-modules, a task-specific weight is finally defined.
Consequently, the task-specific weight is utilized to produce task-adaptive feature maps.

Our main contributions are summarized as follows:
\begin{itemize}
\item We propose a novel feature alignment method, TDM, to define the class-wise channel importance, for fine-grained few-shot classification.
\item Our proposed TDM is highly applicable to prior metric-based few-shot classification models.
\item When combined with the recent few-shot classification models, the TDM achieves the state-of-the-art performance in fine-grained few-shot classification task.
\end{itemize}

\section{Related Works}
\label{sec:rel}
\subsection{Few-Shot Classification}
\begin{figure*}[t!]
    \centering
    \includegraphics[width=1.\textwidth]{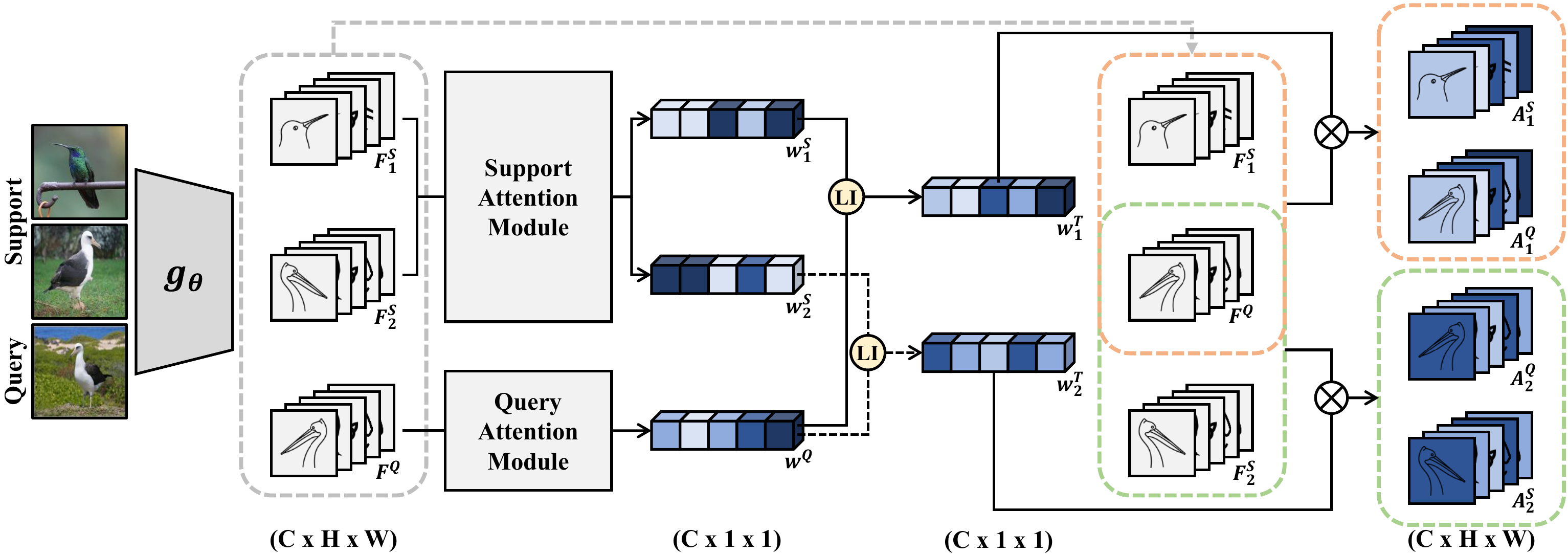}
    \vspace{-0.5cm}
    \caption{
    Overview of our approach. 
    The Task Discrepancy Maximization consists of two sub-modules. 
    Each sub-module takes feature maps $F$ and generates channel weights $w$. 
    The support attention module utilizes feature maps of the support instances as input and finds discriminative channels for each class. 
    Then, it produces a support weight $w^S_i$ for $i$-th category  where the weight holds high values in those channels. 
    On the other hand, the query attention module is fed the query instance and discovers the object-relevant channels of the query. 
    Then, a query weight $w^Q$ from the query attention module emphasizes specific channels with object information about the query. 
    These weights from two sub-modules are integrated by a linear combination to define a task weight $w^T_i$ for each $i$-th category. 
    Finally, task-adaptive feature maps which concentrate on the discriminative regions are obtained by multiplying the task weights to original feature maps.
    }
    \vspace{-0.4cm}
    \label{fig:TDM}
\end{figure*}
The methods of few-shot classification can be divided into two main streams: optimization- and metric-based. 
The concept of optimization-based methods was introduced in MAML~\cite{finn2017model} to learn good initial conditions that can be easily adapted.  
Meta-LSTM~\cite{ravi2016optimization} adopts an LSTM-based meta-learner which is not only for the general initial point but also for effective fine-tuning. 
MetaOptNet~\cite{lee2019meta} employs convex base learners, and provides a differentiation process for end-to-end learning. 
The optimization-based methods show comparable performance, but they need online updates for novel classes.

The metric-based methods learn deep representations by utilizing a predefined\cite{vinyals2016matching, snell2017prototypical, khrulkov2020hyperbolic} or online-trained metric\cite{sung2018learning}. 
MatchNet~\cite{vinyals2016matching} uses an external memory module which augments neural networks and infers categories of the query set by the cosine similarity. 
ProtoNet~\cite{snell2017prototypical} forms prototypes with  a mean feature of each class in support set, and exploits them for computing the distance between a query to each class.
RelationNet~\cite{sung2018learning} utilizes the distance metric learned by a model instead of the predefined metric. 

Metric-based methods generally learn to reduce distances among instances within a class, and we have the same goal since TDM is a module for them. However, TDM enables to compute the distances based on adaptive channel weights by identifying discriminative channels dynamically, while prior techniques treat all the channels equally.

\subsection{Feature Alignment}
Feature alignment methods can be categorized into spatial and channel alignments.
The spatial alignment methods~\cite{hou2019cross, doersch2020crosstransformers, xu2021learning, kang2021relational, wertheimer2021few, zhang2020deepemd} assert that object location differences in the support and query set cause the performance degradation.
CAN~\cite{hou2019cross} computes cross attention maps by calculating correlation for each pair of the classes and query feature maps to highlight the common regions to locate the object.
CTX~\cite{doersch2020crosstransformers} finds a coarse spatial correspondence between the query instance and the support set by the attention\cite{bahdanau2014neural} to produce a query-aligned prototype per each class.
FRN~\cite{wertheimer2021few} reconstructs the feature maps of the support set to the query instance by exploiting a closed-form solution of the ridge regression.

The channel alignment methods~\cite{xu2021learning, kang2021relational, ye2020few, simon2020adaptive} manipulate feature maps to be capable to represent novel classes.
FEAT~\cite{ye2020few} increases the distances among classes of the support set by adopting the transformer~\cite{lin2017structured, vaswani2017attention}.
DMF~\cite{xu2021learning} aligns feature maps of the query instance by the dynamic meta-filter which has both position and channel-specific support knowledge. 
RENet~\cite{kang2021relational} transforms a feature map with self-correlation capturing structural patterns of each image. 

TDM also deals with the feature alignment. 
Unlike existing methods that typically consider a pairwise relationship between the support image and the query image, TDM considers the entire task.  

\section{Method}
\label{sec:met}
The overall architecture of our method is illustrated in \cref{fig:TDM}. 
Given an episode consisting of the support and query instances, feature maps are first computed by the feature extractor. 
However, the feature maps are not optimal for each episode since the feature extractor is trained to find discriminative features for classifying base classes~\cite{ye2020few, zhang2021prototype, roady2020difsim}.
TDM transforms the feature maps by exploiting task-specific weights representing channel-wise discriminative power for a specific task. 
As a result, we aim to focus on the discriminatory details by refining the original feature maps into task-adaptive feature maps. 
In this section, we introduce the components of TDM and their purpose.
First, we formulate the problem in \cref{problem_formulation}.
In~\cref{distance}, we define two representative scores to produce channel weights.
Then, with these scores, we describe two modules of the TDM: SAM and QAM in \cref{support_attention_module} and \cref{query_attention_module}, respectively.
Finally, TDM is described in \cref{task_dircrepancy_maximization} with the discussion in \cref{discussion}.

\subsection{Problem Formulation}\label{problem_formulation}
In standard few-shot classification, we are given two datasets: meta-train set $D_{base}=\left\{\left(x_i,y_i\right),y_i\in{C_{base}}\right\}$ and meta-test set $D_{novel}=\left\{\left(x_i,y_i\right),y_i\in{C_{novel}}\right\}$. 
$C_{base}$ and $C_{novel}$ represent base classes and novel classes, respectively, where they do not overlap (${C_{base}}\cap {C_{novel}}=\phi$). 
Generally, training and testing of few-shot classification are composed of episodes. 
Each episode consists of randomly sampled $N$ classes and each class is composed of $K$ labeled images and $U$ unlabeled images, \textit{i.e.}, $N$-way $K$-shot episode. 
The labeled images are called the support set $S=\left\{\left(x_j, y_j\right)\right\}^{{N}\times{K}}_{j=1}$,
and the unlabeled images are named the query set $Q=\left\{\left(x_j, y_j\right)\right\}_{j=1}^{{N}\times{U}}$, while two sets are disjoint (${S}\cap{Q}=\phi$). The support and query sets are utilized for learning and testing, respectively.

\subsection{Channel-wise Representativeness Scores}\label{distance}
For each pair of $i$-th class and $c$-th channel, we define two channel-wise representativeness scores; intra score $R^{\text{intra}}_{i,c}$, and inter score $R^{\text{inter}}_{i,c}$.
Prior to explaining scores, we first define feature maps $F$ of the support and query instance as follows:
\begin{equation}
    \begin{split}
        \label{feature_maps}
        F^S_{i,j}=g_{\theta}(x^S_{i,j})
        \\
        F^Q=g_{\theta}(x^Q),
    \end{split}
\end{equation}
where $x^S_{i,j}$ is $j$-th instance of $i$-th class in the support set, $x^Q$ is the query instance, and $g_\theta$ is our feature extractor parameterized by $\theta$. 
Each feature map $F$ $\in$ $\mathbb{R}^{{C}\times{H}\times{W}}$ where $C,H,W$ denote the number of channels, height, and width, respectively. 
Additionally, we utilize a prototype\cite{snell2017prototypical} as the representative of each class.
The prototype $F^P_i$ for $i$-th class is computed as follows:
\begin{equation}
    \begin{split}
        \label{prototype}
        F^P_i=\frac{1}{K}\sum^{K}_{j=1}F^S_{i,j}.
    \end{split}
\end{equation}
For each class, we first compute a mean spatial feature to represent a salient object regions. When the $c$-th channel of the prototype for $i$-th class is denoted as ${f^P_{i,c}}\in{\mathbb{R}^{{H}\times{W}}}$, the corresponding mean spatial feature $M^P_i$ is computed as follows:
\begin{equation}
    \begin{split}
        \label{mean_spatial_feature}
        M^P_i=\frac{1}{C}\sum^{C}_{j=1}f^P_{i,j}.
    \end{split}
\end{equation}
Based on this, we further compute the channel-wise representativeness score defined within a class, $R^{\text{intra}}_{i,c}$, for $c$-th channel of $i$-th class as follows:
\begin{equation}
    \begin{split}
        \label{in_class_distance}
        R^{\text{intra}}_{i,c} = \frac{1}{H\times W}\parallel f^P_{i,c} - M^P_i \parallel^2.
    \end{split}
\end{equation}
The score indicates how well the highly activated regions on the channel are matched with the class-wise salient area represented by the mean spatial feature.
On the other hand, the channel-wise representativeness score across classes, $R^{\text{inter}}_{i,c}$, for $c$-th channel of $i$-th class is defined as follows:
\begin{equation}
    \begin{split}
        \label{out_class_distance}
        R^{\text{inter}}_{i,c} = \frac{1}{H \times W}\min_{ 1 \leq j \leq N, j \neq{i} } \parallel f^P_{i,c} - M^P_j \parallel^2.
    \end{split}
\end{equation}
It represents how much $c$-th channel contains the distinct information of each class.
Intuitively, the channel is more discriminative when it has the smaller intra score and the larger inter score.
We utilize both scores to define channel weights.

\subsection{Support Attention Module (SAM)}\label{support_attention_module}
\begin{figure}[H]
    \centering
    \includegraphics[width=\columnwidth]{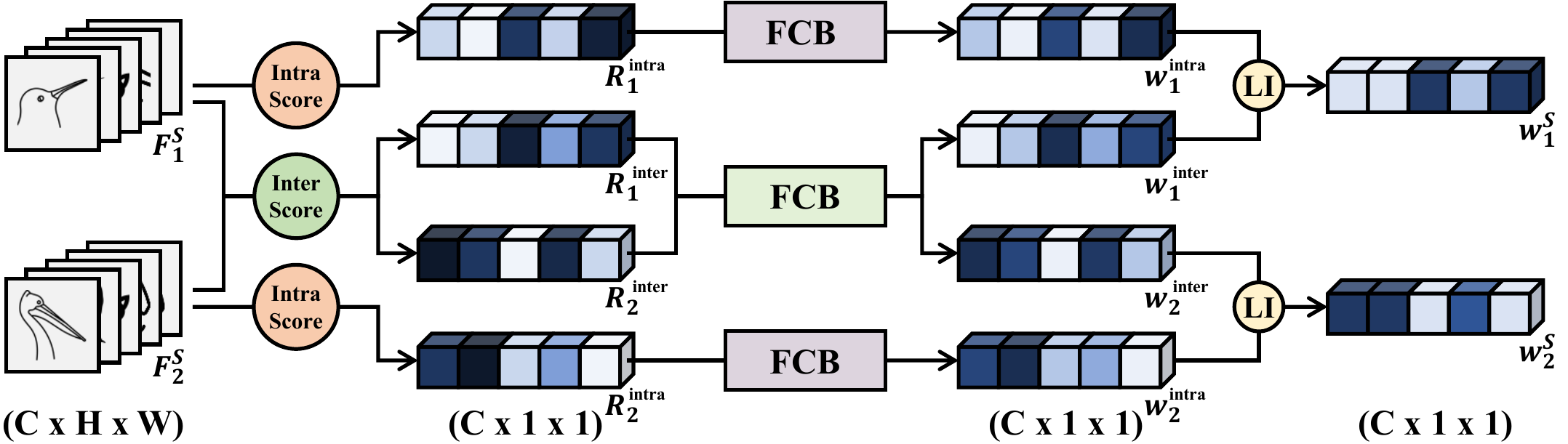}
    \vspace{-0.5cm}
    \caption{
    Schemetic illustration of Support Attention Module.
    }
     \vspace{-0.4cm}
    \label{fig:SAM}
\end{figure}
For each class, our support attention module (SAM) takes the class prototypes as input, and first compute two channel-wise representativeness scores based on \cref{in_class_distance} and \cref{out_class_distance}.
We transform those two scores to two weights, $w^{\text{intra}}_i$ and $w^{\text{inter}}_i$, for $i$-th class as follows:
\begin{equation}
    \begin{split}
        \label{in_out_weight_vector}
        w^{\text{intra}}_i&=b^{\text{intra}}\left(R^{\text{intra}}_{i}\right)
        \\
        w^{\text{inter}}_i&=b^{\text{inter}}\left(R^{\text{inter}}_{i}\right),
    \end{split}
\end{equation}
where $b^{\text{intra}}$ and $b^{\text{inter}}$ are different fully-connected blocks. The architectures of the blocks are reported in \cref{fully_connectec_block}.

The support weight vector $w^{S}_i$ for $i$-th class is defined by a linear combination of the corresponding two weights, $w^{\text{intra}}_i$ and $w^{\text{inter}}_i$, with a balancing hyperparameter $\alpha$ as follows:
\begin{equation}
    \begin{split}
        \label{support_weight_vector}
        w^{S}_i={\alpha}{w^{\text{intra}}_i}+{\left(1-\alpha\right)}{w^{\text{inter}}_i},\: {\alpha}\in{\left[0,1\right]}.
    \end{split}
\end{equation}
The support weight vector for $i$-th class emphasizes discriminative channels of $i$-th class while suppressing channels that corresponds to the common information shared throughout classes in the episode. 
When we multiply the support weight vector for $i$-th class, $w^{S}_i$, to the feature maps, the instances of $i$-th class should be gathered, while other class instances become separated from the $i$-th class.

\subsection{Query Attention Module (QAM)}\label{query_attention_module}
\begin{figure}[h]
    \centering
    \includegraphics[width=\columnwidth]{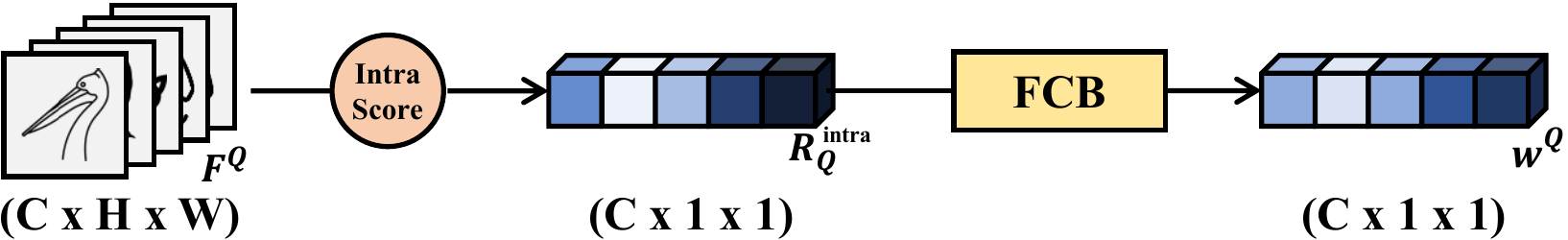}
    \vspace{-0.5cm}
    \caption{Schemetic illustration of Query Attention Module}
    \vspace{-0.3cm}
    \label{fig:QAM}
\end{figure}
Although the support weight vectors are useful to distinguish the class-wise informative channels, we are also encouraged to utilize the query set to overcome the data scarcity in the few-shot learning.
To utilize the query set for the complementary benefits with SAM, we propose the query attention module (QAM).
Since we do not have label information for a query instance, QAM only exploits the relationship among channels within an instance.  
By utilizing the mean spatial feature $M^Q$ defined by the channel-wise mean of query's feature map $F^Q$, we compute the channel-wise representativeness score of query instance, $R^\text{intra}_Q$, for $c$-th channel as follows:
\begin{equation}
    \begin{split}
        \label{qwer}
        R^{\text{intra}}_{Q} = \frac{1}{H\times W}\parallel f^Q_{c} - M^Q \parallel^2,
    \end{split}
\end{equation}
where $f^Q_c$ denotes $c$-th channel of query's feature map.
Then, the query weight $w^{Q}$ is produced by passing the intra score $R^{\text{intra}}_Q$ to the fully connected block $b^Q$ as described in \cref{fully_connectec_block}.
\begin{equation}
    \begin{split}
        \label{query_weight_vector}
        w^{Q}=b^{\text{Q}}\left(R^{\text{intra}}_{Q}\right).
    \end{split}
\end{equation}
The query weight vector highlights object-relevant channels of the query instance while restraining others. 
Therefore, query weight vector assists the model to focus on object-related information.

\begin{table}[t!]
    \centering
    {\small
		\begin{tabular}{|c|c|}
		    \hline
		    \multicolumn{2}{|c|}{\textbf{Fully Connected Block}}\\
		    \hline
		    Layer & Output Size \\
		    \hline
            Input & B $\times$ C \\
            \hline
            Fully Connected Layer & B $\times$ 2C\\
            \hline
            Batch Normalization & B $\times$ 2C\\
            \hline
            ReLU & B $\times$ 2C\\
            \hline
            Fully Connected Layer & B $\times$ C\\
            \hline
            1 + Tanh & B $\times$ C\\
            \hline
		\end{tabular}
	}
	\vspace{-0.1cm}
	\caption{
	The architecture of the fully connected blocks.
    When used in the support attention module, the batch size B is the number of categories which is comprising an episode.
	In the query attention module, B is the number of queries.
	}
	\label{fully_connectec_block}
	\vspace{-0.3cm}
\end{table}

\subsection{Task Discrepancy Maximization (TDM)}\label{task_dircrepancy_maximization}
As the support and query weight vectors computed by SAM and QAM are complementary in their purposes, we use them to produce a task weight vector.
Specifically, the task weight vector $w^T_i$ for $i$-th class is defined by a linear combination of the corresponding support and query weight vectors, $w^S_i$ and $w^Q$, with a hyperparameter $\beta$ as follows:
\begin{equation}
    \begin{split}
        \label{task_weight_vector}
        w^{T}_i={\beta} {w^{S}_i}+{\left(1-\beta\right)} {w^{Q}},\quad {\beta}\in{\left[0,1\right]}.
    \end{split}
\end{equation}
Then, the feature maps of all the support and query instances are transformed with the task weight vector.
Specifically, each feature map $F \in \mathbb{R}^{{C}\times{H}\times{W}}$ is transformed to a task-adaptive feature map $A$ with channel-wise scaling by the task weight vector $w^T_i \in \mathbb{R}^{{C}}$ as follows:
\begin{equation}
    \small
    \begin{split}
        \label{transformation}
        A=w^T_i \odot F =  \left[{w^T_{i,1}}{f_1},{w^T_{i,2}}{f_2}, ... , {w^T_{i,C}}{f_C} \right],
    \end{split}
\end{equation}
where $w^T_{i,j}$ is a scalar value at $j$-th dimension of the vector $w^T_i$, and the $c$-th channel of the feature map $F$ is denoted by ${f_{c}}\in{\mathbb{R}^{{H}\times{W}}}$.

The feature maps of the support instances of $i$-th class are transformed by its corresponding task weight vector $w^T_i$. 
However, since the label of the query is not available, we apply the task weight vector for $i$-th class, $w^T_i$, to the query, when we are testing the query for the $i$-th class.
As a result, based on \cref{transformation}, we compute task adaptive feature maps of the support instances $A^S_{i,j}$ and the query instance $A^Q_i$ transformed by the task weight vector $w^T_i$ as follows:
\begin{equation}
    \begin{split}
        \label{adaptive_feature_map_class}
        A^S_{i,j}&={w^T_i}\odot{F^S_{i,j}}
        \\
        A^Q_i&={w^T_i}\odot{F^Q},
    \end{split}
\end{equation}
where $i$, $j$ denote the class index and instance index within the class, respectively.

For instance, when TDM is applied to the ProtoNet\cite{snell2017prototypical}, the inference process is done by the following criteria:
\begin{equation}
    \begin{split}
        \label{Inferring_class}
        p_{\theta}(y=i|x)=\frac{{\exp}(-d(A^S_i,A^Q_i))}{\sum_{j=1}^{N}{\exp}(-d(A^S_j,A^Q_j))},
    \end{split}
\end{equation}
where $d$ indicates the distance metric, and $A^S_i$ is the prototype computed by the adaptive feature maps of support instances in $i$-th class.

\subsection{Discussion}\label{discussion}
\begin{figure}[t!]
    \centering
    \includegraphics[width=\columnwidth]{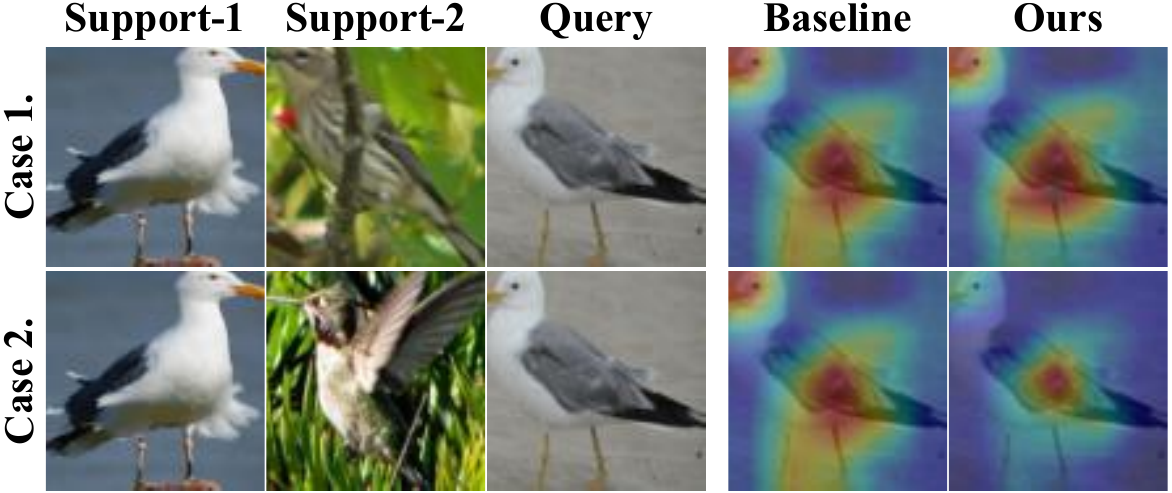}
    \caption{
    2D-aggregated feature activation on 2-way 1-shot.
    (Case 1) If beaks and wings are not similar between species, TDM regards both beaks and wings to be discriminative.
    (Case 2) However, when birds have similar beaks, TDM considers only wings as a discriminative part.
    }
    \vspace{-0.1cm}
    \label{fig:channel_attention_map}
    \vspace{-0.2cm}
\end{figure}
Commonly, it is known that the feature map containing various information about the object is beneficial for a general dataset\cite{huang2020self, li2021learning, li2021beyond}.
However, in a fine-grained dataset, those feature maps are often harmful to infer the categories.
Instead, it is advantageous to focus only on the discriminative parts since the categories share common overall apperance~\cite{ge2019weakly, liu2020filtration, ding2019selective, zheng2019looking}.
Moreover, unlike general fine-grained classification where the discriminative region of each category is almost constant, the distinct region of each class in few-shot setting may vary depending on the contents of the episode. 
Therefore, the key point of fine-grained few-shot classification is dynamically discovering the discriminative regions of each class based on very small number of instances.
In \cref{fig:channel_attention_map}, we observe that the baseline model treats all characteristics equally regardless of the composition of each episode.
However, TDM adaptively allows the model to highlight regions expected as discriminative parts and suppress other regions.
Thus, TDM is a specialized module for fine-grained few-shot classification.

\section{Experiments}
\label{sec:exp}
In this section, we evaluate TDM on standard fine-grained classification benchmarks. To verify high adaptability of TDM, we apply it to various existing methods: ProtoNet\cite{snell2017prototypical},  DSN\cite{simon2020adaptive}, CTX\cite{doersch2020crosstransformers}, and FRN\cite{wertheimer2021few}.
For a fair comparison, we reproduce each baseline model with the same hyperparameter and implementation details regardless of whether TDM is attached or not. 
In each table, $\dagger$ indicates the reproduced version of the baseline model. 
While TDM generally exploits the prototype\cite{snell2017prototypical} defined in \cref{prototype} for computing the intra and inter score, in CTX, it utilizes a query-aligned prototype\cite{doersch2020crosstransformers}.  

\subsection{Datasets}
\begin{table}[t]
    \centering
    {\small
		\begin{tabular}{l | c c c c}
		    \hlineB{2.5}
		    \multicolumn{1}{l}{\textbf{Dataset}} & \textbf{$C_{all}$} & \textbf{$C_{train}$} & \textbf{$C_{val}$} & \textbf{$C_{test}$} \\
		    \hlineB{2.5}
            \multicolumn{1}{l}{CUB-200-2011} & 200 & 100 & 50 & 50 \\
            \multicolumn{1}{l}{Aircraft} & 100 & 50 & 25 & 25 \\
            \multicolumn{1}{l}{meta-iNat} & 1135 & 908 & - & 227 \\
            \multicolumn{1}{l}{tiered meta-iNat} & 1135 & 781 & - & 354 \\
            \multicolumn{1}{l}{Stanford-Cars} & 196 & 130 & 17 & 49 \\
            \multicolumn{1}{l}{Stanford-Dogs} & 120 & 70 & 20 & 30 \\
            \multicolumn{1}{l}{Oxford-Pets} & 37 & 20 & 7 & 10 \\
            \hlineB{2.5}
		\end{tabular}
	}
    \vspace{-0.1cm}
	\caption{
	The splits of datasets. While $C_{all}$ is the number of total classes, $C_{train}$, $C_{val}$, $C_{test}$ are the number of training, validation, and test classes, respectively.
	The classes of subsets are disjoint.
	}
	\label{dataset_split}
	\vspace{-0.4cm}
\end{table}
We utilize seven benchmarks for few-shot classification: CUB-200-2011, Aircraft, meta-iNat, tiered meta-iNat, Stanford-Cars, Stanford-Dogs, and Oxford-Pets. 
The data split for each dataset is provided in \cref{dataset_split}.

\noindent\textbf{CUB-200-2011}\cite{wah2011caltech} is an image dataset with 11,788 photos of 200 bird species. 
This benchmark can be used in two ways: raw form\cite{chen2019closer} or preprocessed form by a human-annotated bounding box\cite{ye2020few, zhang2020deepemd}. 
For a fair comparison, we utilizes both settings. 
Following \cite{chen2019closer}, we split this benchmark and our split is the same with \cite{wertheimer2021few}.

\noindent\textbf {Aircraft}\cite{maji2013fine} contains 10,000 airplane images of 100 models. 
The main challenge of this benchmark is the similarity by the symbol of the airline. 
Although the types of aircraft are different, the symbol can be equivalent when they belong to the same airline, making it more difficult. 
Following \cite{wertheimer2021few}, we preprocess all images of this benchmark based on the bounding box and divide the dataset.

\noindent\textbf{meta-iNat}\cite{wertheimer2019few, van2018inaturalist} is a realistic, heavy-tailed benchmark for few-shot classification. 
It contains 1,135 animal species spanning 13 super categories, and the number of images in each class is imbalanced with a range between 50 and 1000 images.
We follow the dataset split introduced in \cite{wertheimer2019few}. 
While a full 227-way evaluation scheme, that each episode consists of all testing categories at once, is adopted in \cite{wertheimer2019few}, we employ a standard 5-way few-shot evaluation scheme following \cite{wertheimer2021few}.

\noindent\textbf{tiered meta-iNat}\cite{wertheimer2019few} is comprised of the same images with meta-iNat.
However, this benchmark divides the dataset by super categories.
The super categories of test dataset are Insects and Arachnids, while the super categories of train dataset are Plant, Bird, Mammal, and etc. 
Therefore, a large domain gap exists between the train and the test classes. 
As following \cite{wertheimer2021few}, we also employ a standard 5-way few-shot evaluation scheme in this benchmark.

\noindent\textbf{Stanford Cars}\cite{krause20133d} contains 16,185 images of 196 classes of cars. 
Classes are typically at the level of Year, Brand and Model name, e.g., 2012 Tesla Model S and 2012 BMW M3 coupe. 
This dataset was introduced by \cite{li2019revisiting} for the few-shot classification.
Likewise, we adopt the same data split with \cite{li2019revisiting}.

\noindent\textbf{Stanford Dogs}\cite{khosla2011novel} is also introduced by \cite{li2019revisiting} for fine-grained few-shot classification. This dataset contains 20,580 images of 120 breeds of dogs around the world. Our split is the same with \cite{li2019revisiting}.

\noindent\textbf{Oxford Pets}\cite{parkhi2012cats} is comprised of 37 pet categories with roughly 200 images for each class. 
Since the lack of training images generally causes overfitting, the generalization capability is essential to attain high accuracy in this benchmark.
To our knowledge, this benchmark dataset has never been used for few-shot classification. 
Therefore, we randomly divide this dataset by referring to the split ratio of the other datasets.
We report the split information in the supplementary material.

\begin{table}[t]
    \centering
    {\small
		\begin{tabular}{l | c c c c}
		    \hlineB{2.5}
		    \multicolumn{1}{l}{\multirow{2}{*}{\textbf{Model}}} & \multicolumn{2}{c}{\textbf{Conv-4}} & \multicolumn{2}{c}{\textbf{ResNet-12}} \\
		    \multicolumn{1}{c}{}& \textbf{1-shot} & \textbf{5-shot} & \textbf{1-shot} & \textbf{5-shot} \\
		    \hlineB{2.5}
		    \multicolumn{1}{l}{MatchNet\cite{vinyals2016matching, ye2020few, zhang2020deepemd}} & 67.73 & 79.00 & 71.87 & 85.08 \\
		    \multicolumn{1}{l}{ProtoNet\cite{snell2017prototypical, ye2020few, zhang2020deepemd}} & 63.73 & 81.50 & 66.09 & 82.50 \\
		    \multicolumn{1}{l}{FEAT{$^{\ast}$}\cite{ye2020few}} & 68.87 & 82.90 & 73.27 & 85.77 \\
		    \multicolumn{1}{l}{DeepEMD\cite{zhang2020deepemd}} & - & - & 75.65 & 88.69 \\
		    \multicolumn{1}{l}{RENet\cite{kang2021relational}} & - & - & 79.49 & 91.11 \\
		    \hlineB{1.0}
            \multicolumn{1}{l}{ProtoNet{$^{\dagger}$}\cite{snell2017prototypical}} & 62.90 & 84.13 & 78.99 & 90.74 \\
            \multicolumn{1}{l}{~~~+ TDM} & 69.94 & 86.96 & 79.58 & 91.28 \\
            \hlineB{1.}
            \multicolumn{1}{l}{DSN{$^{\dagger}$}\cite{simon2020adaptive}} & 72.09 & 85.03 & 80.51 & 90.23 \\
            \multicolumn{1}{l}{~~~+ TDM} & 73.38 & 86.07 & 81.33 & 90.65 \\
                        \hlineB{1.}
            \multicolumn{1}{l}{CTX{$^{\dagger}$}\cite{doersch2020crosstransformers}} & 72.14 & 87.23 & 80.67 & 91.55 \\
            \multicolumn{1}{l}{~~~+ TDM} & \textbf{74.68} & 88.36 & 83.28 & 92.74 \\
                        \hlineB{1.}
            \multicolumn{1}{l}{FRN{$^{\dagger}$}\cite{wertheimer2021few}} & 73.24 & 88.33 & 83.16 & 92.42 \\
            \multicolumn{1}{l}{~~~+ TDM} & 74.39 & \textbf{88.89} & \textbf{83.36} & \textbf{92.80} \\
            \hlineB{2.5}
		\end{tabular}
	}
	\vspace{-0.1cm}
	\caption{Performance on CUB using bounding-box cropped images as input. ``$\ast$" denotes reproduced one in RENet. Confidence intervals for our implemented model are all below 0.23.}
	\label{CUB_cropped}
	\vspace{-0.4cm}
\end{table}
\subsection{Implementation Details}
\begingroup
\setlength{\tabcolsep}{4pt} 
\renewcommand{\arraystretch}{1.0} 
\begin{table}[t]
    \centering
    {\small
		\begin{tabular}{l | c c c}
		    \hlineB{2.5}
		    \multicolumn{1}{l}{\textbf{Model}} & \textbf{Backbone} & \textbf{1-shot} & \textbf{5-shot} \\
		    \hlineB{2.5}
            \multicolumn{1}{l}{Baseline{$^{\flat}$}\cite{chen2019closer}} & ResNet-18 & 65.51$\pm$0.87 & 82.85$\pm$0.55 \\
            \multicolumn{1}{l}{Baseline++{$^{\flat}$}\cite{chen2019closer}} & ResNet-18 & 67.02$\pm$0.90 & 83.58$\pm$0.54 \\
            \multicolumn{1}{l}{MatchNet{$^{\flat}$}\cite{chen2019closer, vinyals2016matching}} & ResNet-18 & 73.42$\pm$0.89 & 84.45$\pm$0.58 \\
            \multicolumn{1}{l}{ProtoNet{$^{\flat}$}\cite{chen2019closer, snell2017prototypical}} & ResNet-18 & 72.99$\pm$0.88 & 86.65$\pm$0.51 \\
            \multicolumn{1}{l}{MAML{$^{\flat}$}\cite{chen2019closer, finn2017model}} & ResNet-18 & 68.42$\pm$1.07 & 83.47$\pm$0.62 \\
            \multicolumn{1}{l}{RelatioNet{$^{\flat}$}\cite{chen2019closer, sung2018learning}} & ResNet-18 & 68.58$\pm$0.94 & 84.05$\pm$0.56 \\
            \multicolumn{1}{l}{S2M2{$^{\flat}$}\cite{mangla2020charting}} & ResNet-18 & 71.43$\pm$0.28 & 85.55$\pm$0.52 \\
            \multicolumn{1}{l}{Neg-Cosine{$^{\flat}$}\cite{liu2020negative}} & ResNet-18 & 72.66$\pm$0.85 & 89.40$\pm$0.43 \\
            \multicolumn{1}{l}{Afrasiyabi \textit{et al.}{$^{\flat}$}\cite{afrasiyabi2020associative}} & ResNet-18 & 74.22$\pm$1.09 & 88.65$\pm$0.55 \\
            \hlineB{1.}
            \multicolumn{1}{l}{ProtoNet{$^{\dagger}$}\cite{snell2017prototypical}} & ResNet-12 & 78.58$\pm$0.22 & 89.83$\pm$0.12 \\
            \multicolumn{1}{l}{~~~+ TDM} & ResNet-12 & 79.11$\pm$0.22 & 90.83$\pm$0.11\\
            \hlineB{1.}
            \multicolumn{1}{l}{DSN{$^{\dagger}$}\cite{simon2020adaptive}} & ResNet-12 & 80.47$\pm$0.20 & 89.92$\pm$0.12 \\
            \multicolumn{1}{l}{~~~+ TDM} & ResNet-12 & 80.58$\pm$0.20 & 89.95$\pm$0.12\\
            \hlineB{1.}
            \multicolumn{1}{l}{CTX{$^{\dagger}$}\cite{doersch2020crosstransformers}} & ResNet-12 & 80.95$\pm$0.21 & 91.54$\pm$0.11\\
            \multicolumn{1}{l}{~~~+ TDM} & ResNet-12 & 83.45$\pm$0.19 & 92.49$\pm$0.11\\
            \hlineB{1.}
            \multicolumn{1}{l}{FRN{$^{\dagger}$}\cite{wertheimer2021few}} & ResNet-12 & 83.54$\pm$0.19 & 92.96$\pm$0.10\\
            \multicolumn{1}{l}{~~~+ TDM} & ResNet-12 & \textbf{84.36$\pm$0.19} & \textbf{93.37$\pm$0.10}\\
            \hlineB{2.5}
		\end{tabular}
	}
	\vspace{-0.1cm}
	\caption{Performance on CUB using raw images as input. ``$\flat$" denotes larger backbones than \textit{ResNet-12}.}
	\label{CUB_raw}
	\vspace{-0.4cm}
\end{table}
\endgroup
\noindent\textbf{Architecture.} 
We adopt common protocols from recent few-shot classification works\cite{kim2019edge, chen2021pareto, zhao2021looking, hong2021reinforced, zhang2021rethinking}; we employ Conv-4 and ResNet-12.
While both backbone networks accept an image of size 84$\times$84, the size of feature maps is different according to the backbone network.
ResNet-12 yields a feature map with size 640$\times$5$\times$5 while Conv-4 offers 64$\times$5$\times$5 shape of the feature map.
For our proposed TDM, we additionally utilize fully-connected layer blocks where the size of blocks are proportional to the dimension of channels of the feature maps as described in \cref{fully_connectec_block}. 
\newline
\textbf{Training Details.} 
Following existing methods\cite{chen2019closer, wang2019simpleshot, ye2020few, zhang2020deepemd, wertheimer2021few}, we use standard data augmentation techniques including random crop, horizontal flip, and color jitter. 
The $\alpha, \beta$ in \cref{support_weight_vector}, \cref{task_weight_vector} are fixed to 0.5 and other parameters are adopted from \cite{wertheimer2021few} which is our baseline model. 
More details are described in the supplementary material.
To prevent overfitting, we add random noise between -0.2 and 0.2 to the task weight for each category from TDM.
\newline
\textbf{Evaluation Details.} 
For the N-way K-shot, we conduct few-shot classification on 10,000 randomly sampled episodes where it contains 16 queries per class. 
We report average classification accuracy with 95\% confidence intervals.
\begin{figure*}
    \centering
    \begin{subfigure}[b]{0.33\textwidth}
        \centering
        \includegraphics[width=\textwidth]{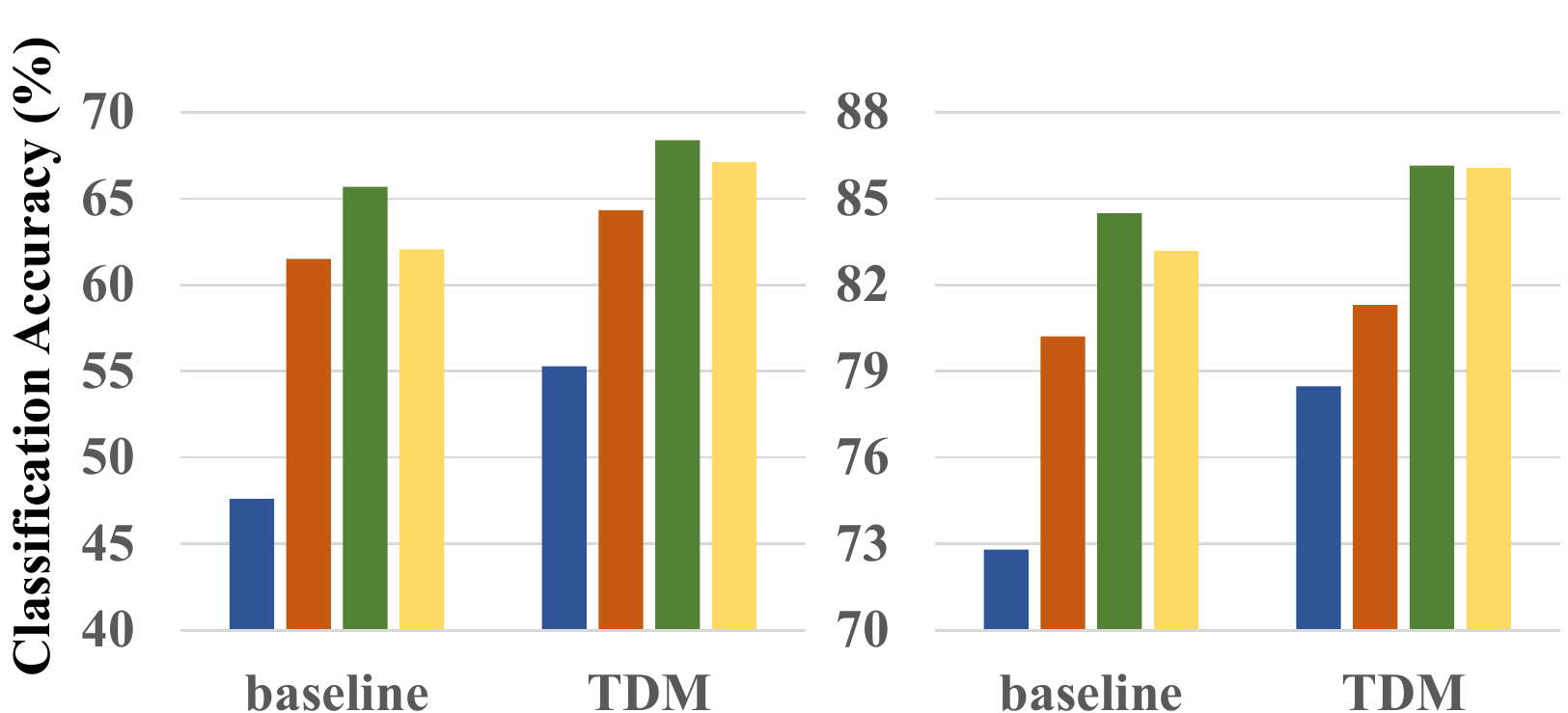}
        \caption{Stanford Cars}
        \vspace{-0.3cm}
        \label{fig:y equals x}
    \end{subfigure}
    \hfill
    \begin{subfigure}[b]{0.33\textwidth}
        \centering
        \includegraphics[width=\textwidth]{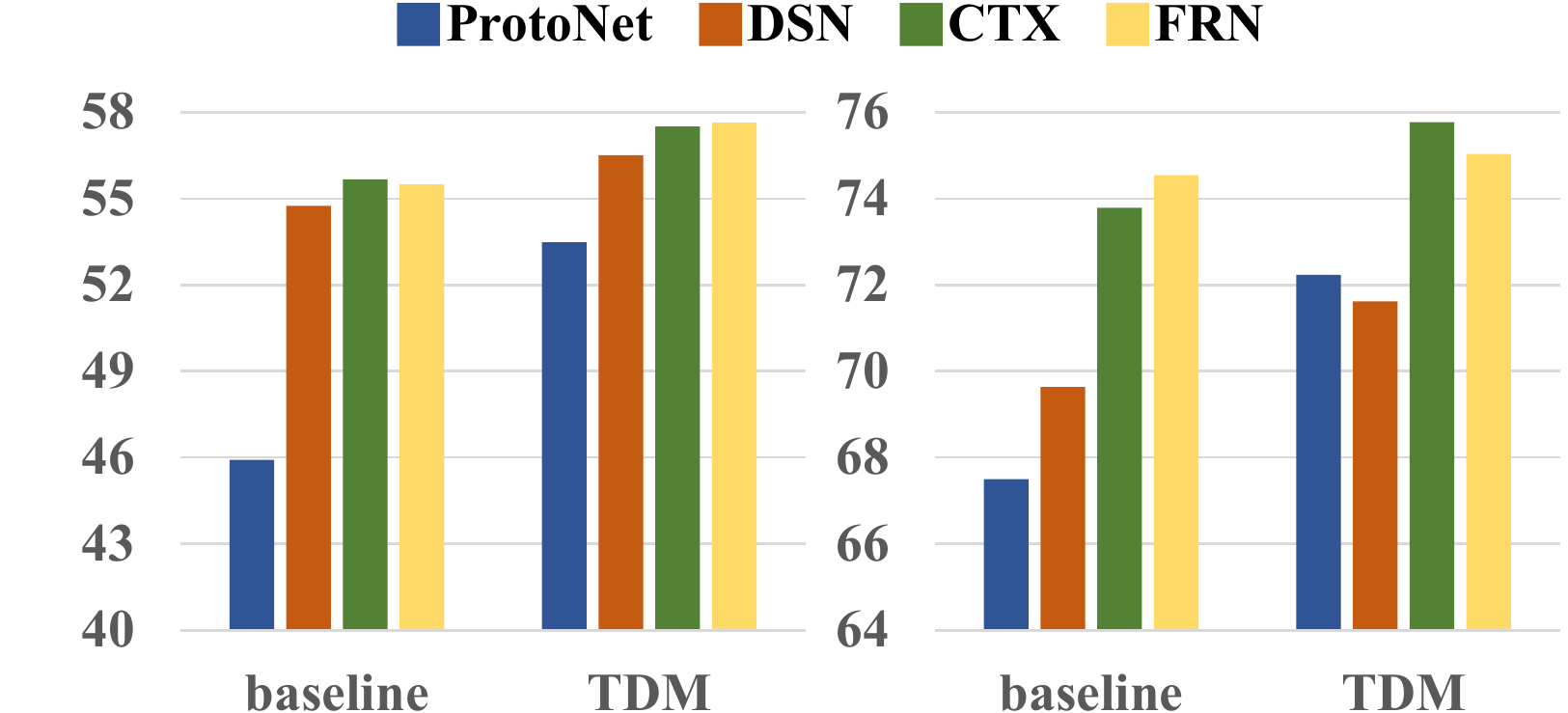}
        \caption{Stanford Dogs}
        \vspace{-0.3cm}
        \label{fig:three sin x}
    \end{subfigure}
    \hfill
        \begin{subfigure}[b]{0.33\textwidth}
        \centering
        \includegraphics[width=\textwidth]{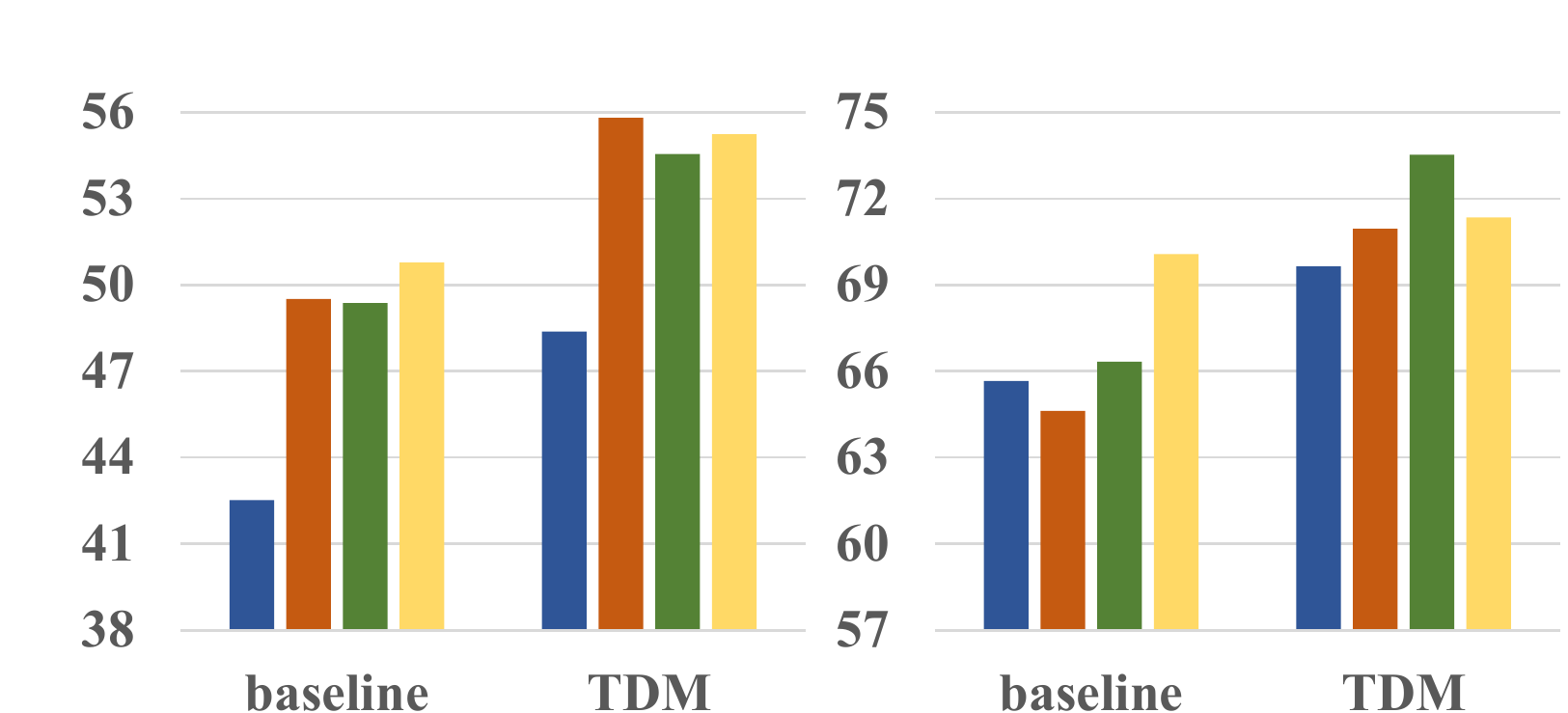}
        \caption{Oxford Pets}
        \vspace{-0.3cm}
        \label{fig:three sin x}
    \end{subfigure}
    \caption{
    Accuracies on additional datasets. 
    The left and right graphs for each dataset show 1-shot accuracies and 5-shot accuracies, respectively.
    The left side of the graph reports the performance of baseline models, while the right side shows performance with TDM. The baseline methods are differentiated with colors.
    }
    \label{fig:additional_dataset}
    \vspace{-0.4cm}
\end{figure*}

\subsection{Comparison to Existing Methods}
\noindent\textbf{CUB-200-2011 results.} \cref{CUB_cropped} and \cref{CUB_raw} reports the results of TDM and baseline few-shot classification methods. 
Although some cases do not surpass confidence intervals, our TDM consistently improves the performance of baselines models in all cases, and achieves state-of-the-art scores regardless the depth of the backbone network.
Especially, TDM improves more than 7\% on ProtoNet in the 1-shot scenario of CUB with Conv-4. 
\newline
\textbf {Aircraft results.} 
As shown in \cref{aircraft}, TDM improves the performances of the baseline models in all cases. 
The increases are beyond the confidence intervals regardless of the type of baseline models and the number of labeled images. 
As a result, we attain top accuracy scores for all benchmarks by large margin. Specifically, TDM boosts the performance of CTX\cite{doersch2020crosstransformers} up to 7\% on the Conv-4 network.
These results demonstrates the effectiveness of TDM.
\newline
\textbf{meta-iNat results.} 
We validate the generalization ability of TDM in this benchmark. 
Meta-iNat is vulnerable to overfitting since the validation set does not exist.
However, as shown in \cref{inat}, TDM is not only robust to overfitting issues but also powerful in generalization capability.
Consequently, TDM shows consistent improvements over the baselines, particularly on ProtoNet where TDM assists them to retain competitive results compared to state-of-the-art methods.
\newline
\textbf{tiered meta-iNat results.} 
We further validate the generalization capability of TDM in a more difficult configuration where the super categories of train set and test set do not overlap. 
Specifically, TDM enhances the performance in most configurations and also accomplishes the best performance in the 1-shot scenario with FRN. 
For the slight decrease in a 5-shot scenario, $\lambda$, the learnable parameter in FRN, is responsible. 
In general, large $\lambda$ shows good performance when a domain gap exists but TDM restrains the $\lambda$ to be relatively small.
We think that this is because TDM assists the classifier to focus on discriminative channels\cite{wertheimer2021few}.
\newline
\textbf{Stanford Cars, Stanford Dogs and Oxford Pets results.} 
Unlike previous benchmarks, these datasets were not evaluated in our baseline models\cite{snell2017prototypical, simon2020adaptive, doersch2020crosstransformers ,wertheimer2021few}. 
To further validate the effectiveness of TDM, we additional conduct experiments on those fine-grained datasets with Conv-4.
As reported in \cref{fig:additional_dataset}, TDM is capable to improve the performance that the confidence interval does not overlap in all cases regardless of the baseline models.
In detail, TDM shows performance boosts in which their accuracies are 4.44 and 3.27 points higher than the baseline at the 1-shot and 5-shot scenarios, respectively.

Throughout the extensive experiments on seven benchmark datasets, we clearly validate the strength of TDM in fine-grained few-shot classification. 
To summarize, we improve the performance of the baseline models in all benchmarks datasets, achieving the state-of-the-art results except the one case: FRN on the tiered meta-iNat 5-shot scenario.

\begingroup
\setlength{\tabcolsep}{6pt} 
\renewcommand{\arraystretch}{1.0} 
\begin{table}[t]
    \centering
    {\small
		\begin{tabular}{l | c c c c}
		    \hlineB{2.5}
		    \multicolumn{1}{l}{\multirow{2}{*}{\textbf{Model}}} & \multicolumn{2}{c}{\textbf{Conv-4}} & \multicolumn{2}{c}{\textbf{ResNet-12}} \\
		    \multicolumn{1}{c}{} & \textbf{1-shot} & \textbf{5-shot} & \textbf{1-shot} & \textbf{5-shot} \\
		    \hlineB{2.5}
            \multicolumn{1}{l}{ProtoNet{$^{\dagger}$}\cite{snell2017prototypical}} & 47.37 & 68.96 & 67.28 & 83.21 \\
            \multicolumn{1}{l}{~~~+ TDM} & 50.55 & 71.12 & 69.12 & \textbf{84.77} \\
            \hlineB{1.}
            \multicolumn{1}{l}{DSN{$^{\dagger}$}\cite{simon2020adaptive}} & 52.22 & 68.75 & 70.23 & 
            83.05 \\
            \multicolumn{1}{l}{~~~+ TDM} & 53.77 & 69.56 & \textbf{71.57} & 83.65 \\
            \hlineB{1.}
            \multicolumn{1}{l}{CTX{$^{\dagger}$}\cite{doersch2020crosstransformers}} & 51.58 & 68.12 & 65.53 & 79.31 \\
            \multicolumn{1}{l}{~~~+ TDM} & \textbf{55.15} & 70.45 & 69.42 & 83.25 \\
            \hlineB{1.}
            \multicolumn{1}{l}{FRN{$^{\dagger}$}\cite{wertheimer2021few}} & 53.12 & 70.84 & 69.58 & 
            82.98 \\
            \multicolumn{1}{l}{~~~+ TDM} & 54.21 & \textbf{71.37} & 70.89 & 84.54 \\
            \hlineB{2.5}
		\end{tabular}
	}
	\vspace{-0.1cm}
	\caption{Performance on Aircraft. Confidence intervals for our implemented model are all below 0.25.}
	\label{aircraft}
	\vspace{-0.4cm}
\end{table}
\endgroup

\begingroup
\setlength{\tabcolsep}{6pt} 
\renewcommand{\arraystretch}{1.0} 
\begin{table}[t]
    \centering
    {\small
		\begin{tabular}{l | c c c c}
		    \hlineB{2.5}
		    \multicolumn{1}{l}{\multirow{2}{*}{\textbf{Model}}} & \multicolumn{2}{c}{\textbf{meta-iNat}} & \multicolumn{2}{c}{\textbf{tiered meta-iNat}} \\
		    \multicolumn{1}{l}{} & \textbf{1-shot} & \textbf{5-shot} & \textbf{1-shot} & \textbf{5-shot} \\
		    \hlineB{2.5}
            \multicolumn{1}{l}{ProtoNet{$^{\dagger}$}\cite{snell2017prototypical}} & 55.37 & 76.30 & 34.41 & 57.60 \\
            \multicolumn{1}{l}{~~~+ TDM} & 61.82 & 79.95 & 38.30 & 61.18 \\
            \hlineB{1.}
            \multicolumn{1}{l}{DSN{$^{\dagger}$}\cite{simon2020adaptive}} & 60.06 & 76.15 & 40.83 & 58.34 \\
            \multicolumn{1}{l}{~~~+ TDM} & 61.87 & 78.07 & 41.00 & 58.66 \\
            \hlineB{1.}
            \multicolumn{1}{l}{CTX{$^{\dagger}$}\cite{doersch2020crosstransformers}} & 60.80 & 78.57 & 42.24 & 60.54\\
            \multicolumn{1}{l}{~~~+ TDM} & 63.26 & 80.75 & 43.90 & 62.29 \\
            \hlineB{1.}
            \multicolumn{1}{l}{FRN{$^{\dagger}$}\cite{wertheimer2021few}} & 61.98 & 80.04 & 43.95 & \textbf{63.45} \\
            \multicolumn{1}{l}{~~~+ TDM} & \textbf{63.97} & \textbf{81.60} & \textbf{44.05} & 62.91 \\
            \hlineB{2.5}
		\end{tabular}
	}
	\vspace{-0.1cm}
	\caption{
	Performance on meta-iNat and tiered meta-iNat using Conv-4 backbones.
	Confidence intervals for our implemented model are all below 0.23.
	}
	\label{inat}
	\vspace{-0.1cm}
\end{table}
\endgroup

\section{Ablation Study}
We conduct ablation study with ProtoNet\cite{snell2017prototypical} on the Conv-4 backbone using CUB\_cropped and Aircraft datasets.
\label{sec:abl}
\vspace{-0.1cm}
\subsection{Effect of Submodules}
\cref{samqam} reports the effects of sub-modules of TDM. 
We observe that both SAM and QAM consistently improve the classification accuracies.
As the second and fourth rows show, SAM improves the baseline by a large gain up to 11\%.
This large gain confirms that recognizing discriminative channels for each category is crucial for fine-grained few shot classification.
Furthermore, although the improvement of QAM is slightly lower than SAM, QAM is shown to be effective for all scenarios.
We think that this is because the object-relevant channels not always represent discriminate channels.
More importantly, the performances can be further boosted when combined. This validates that two sub-modules are complementary to one another.

\begingroup
\setlength{\tabcolsep}{6pt} 
\renewcommand{\arraystretch}{1.0} 
\begin{table}[t]
    \centering
    {\small
		\begin{tabular}{c c c c c c}
		    \hlineB{2.5}
            \multicolumn{1}{c}{\multirow{2}{*}{\textbf{SAM}}} &             \multicolumn{1}{c}{\multirow{2}{*}{\textbf{QAM}}} & \multicolumn{2}{c}{\textbf{CUB\_cropped}} & \multicolumn{2}{c}{\textbf{Aircraft}} 
            \\
		    & & \textbf{1-shot} & \textbf{5-shot} & \textbf{1-shot} & \textbf{5-shot} 
		    \\
		    \hlineB{2.5}
		    - & - & 62.90 & 84.13 & 47.37 & 68.96 \\
		    \cmark & - & 68.53 & 85.95 & 49.45 & 69.33 \\
		    - & \cmark & 65.11 & 84.82 & 48.96 & 70.85 \\
		    \cmark & \cmark & \textbf{69.94} & \textbf{86.96} & \textbf{50.55} & \textbf{71.12} \\
		    \hlineB{2.5}
		\end{tabular}
	}
	\vspace{-0.1cm}
	\caption{Effects of SAM and QAM.}
	\label{samqam}
	\vspace{-0.3cm}
\end{table}
\subsection{Choice of Pooling Function}
We also study the effects of pooling functions, as described in \cref{pooling_function}.
As reported in the second and third rows, both pooling methods boost the performance since they are capable of representing the details of the objects.
However, as shown in \cref{fig:pooling}, max-pooling function has its limitation in that it is vulnerable to noise.
So thus, we adopt the average-pooling function to predict discriminative channels.

\subsection{Metric Compatibility}
Following our baselines~\cite{snell2017prototypical, simon2020adaptive, doersch2020crosstransformers, wertheimer2021few}, we employ the Euclidean distance for TDM. 
Since TDM is compatible with other metrics, we evaluate it with the cosine distance. 
As reported in \cref{cosine_simialrity}, TDM improves the baseline.
This validates that TDM's flexibility to commonly used distance metrics: the Euclidean and cosine distances.

\section{Limitation}
\label{sec:lim}
Since representing the whole objects could be rather harmful for classifying fine-grained categories unlike the general classification task, TDM is developed for highlighting features discriminative for fine-grained details. Thus, the benefits of TDM could be limited in coarse-grained tasks.

\begin{table}[t]
    \centering
    {\small
		\begin{tabular}{c c c c c c}
		    \hlineB{2.5}
		    \multicolumn{1}{c}{\multirow{2}{*}{\textbf{TDM}}} & \multicolumn{1}{c}{\multirow{2}{*}{\textbf{Pooling}}} & \multicolumn{2}{c}{\textbf{CUB\_cropped}} & \multicolumn{2}{c}{\textbf{Aircraft}} 
		    \\
		    \multicolumn{1}{c}{} & \multicolumn{1}{c}{} & \textbf{1-shot} & \textbf{5-shot} & \textbf{1-shot} & \textbf{5-shot} 
		    \\
		    \hlineB{2.5}
		    \multicolumn{1}{c}{-} & \multicolumn{1}{c}{-} & 62.90 & 84.13 & 47.37 & 68.96 \\
            \multicolumn{1}{c}{\cmark} & \multicolumn{1}{c}{Max} & 67.23 & 86.73 & 50.16 & \textbf{71.32} \\
            \multicolumn{1}{c}{\cmark} & \multicolumn{1}{c}{Avg} & \textbf{69.94} & \textbf{86.96} & \textbf{50.55} & 71.12 \\
            \hlineB{2.5}
		\end{tabular}
	}
	\vspace{-0.1cm}
	\caption{Effects of Pooling functions. Top row is the baseline result. 
	}
	\label{pooling_function}
	\vspace{-0.2cm}
\end{table}

\begin{table}[t]
    \centering
    {\small
		\begin{tabular}{c c c c c}
		    \hlineB{2.5}
		    \multicolumn{1}{c}{\multirow{2}{*}{\textbf{TDM}}} & \multicolumn{2}{c}{\textbf{CUB\_cropped}} & \multicolumn{2}{c}{\textbf{Aircraft}} \\
		    \multicolumn{1}{c}{} & \textbf{1-shot} & \textbf{5-shot} & \textbf{1-shot} & \textbf{5-shot} \\
		    \hlineB{2.5}
            \multicolumn{1}{c}{-} & 68.69 & 82.89 & 48.36 & 63.45 \\
            \multicolumn{1}{c}{\cmark} & \textbf{70.47} & \textbf{84.34} & \textbf{49.21} & \textbf{66.26} \\
            \hlineB{2.5}
		\end{tabular}
	}
	\vspace{-0.1cm}
	\caption{Effectiveness of TDM with the cosine distance}
	\label{cosine_simialrity}
	\vspace{-0.4cm}
\end{table}

\section{Conclusion}
\label{sec:con}
In this paper, we introduced Task Discrepancy Maximization (TDM), a tailored module for fine-grained few-shot classification.
TDM produces channel weights that emphasize features of fine discriminative details to distinguish similar classes with two submodules: Support Attention Module (SAM) and Query Attention Module (QAM).
Our extensive experiments on several fine-grained benchmarks validated the merits of our proposed TDM in terms of its effectiveness and high applicability with the prior few-shot classification methods.
As a future direction, we will investigate how the significance for each channel vary in other computer vision tasks and extend this module to those tasks.

\vspace{5pt}
\noindent\textbf{Acknowledgements.} This work was supported in part by MCST/KOCCA (No. R2020070002), MSIT/IITP (No. 2020-0-00973, 2019-0-00421, 2020-0-01821, 2021-0-02003, and 2021-0-02068), and MSIT\&KNPA/KIPoT (Police Lab 2.0, No. 210121M06).

\newpage
\small
\bibliographystyle{ieee_fullname}
\bibliography{egbib}

\end{document}